\pdfoutput=1

\documentclass[11pt]{article}

\usepackage[final]{acl}

\usepackage{times}
\usepackage{latexsym}

\usepackage[T1]{fontenc}

\usepackage[utf8]{inputenc}

\usepackage{microtype}

\usepackage{inconsolata}

\usepackage{booktabs}
\usepackage{amsmath}
\usepackage{amsfonts}
\usepackage{graphicx}
\usepackage{tabularx}
\usepackage{multirow}

\newcommand{\benchmark}{XC-Translate}

\title{Towards Cross-Cultural Machine Translation with\\Retrieval-Augmented Generation from Multilingual Knowledge Graphs}

\author{Simone Conia\thanks{\ Work carried out partially while at Apple. These authors contributed equally.} \\
	Sapienza University of Rome \\
	\texttt{simone.conia@uniroma1.it} \\\And{}
	Daniel Lee\footnotemark[1] \\
	Adobe \\
	\texttt{dlee1@adobe.com} \\\And{}
	Min Li \\
	Apple \\
	\texttt{min\_li6@apple.com} \\\AND{}
	Umar Farooq Minhas \\
	Apple \\
	\texttt{ufminhas@apple.com} \\\And{}
	Saloni Potdar \\
	Apple \\
	\texttt{s\_potdar@apple.com} \\\And{}
	Yunyao Li \\
	Adobe \\
	\texttt{yunyaol@adobe.com} \\
}

\begin{document}
\maketitle
\begin{abstract}
	Translating text that contains entity names is a challenging task, as cultural-related references can vary significantly across languages. These variations may also be caused by \textit{transcreation}, an adaptation process that entails more than \textit{transliteration} and \textit{word-for-word} translation. In this paper, we address the problem of cross-cultural translation on two fronts: (i) we introduce \benchmark{}, the first large-scale, manually-created benchmark for machine translation that focuses on text that contains potentially culturally-nuanced entity names, and (ii) we propose KG-MT, a novel end-to-end method to integrate information from a multilingual knowledge graph into a neural machine translation model by leveraging a dense retrieval mechanism. Our experiments and analyses show that current machine translation systems and large language models still struggle to translate texts containing entity names, whereas KG-MT outperforms state-of-the-art approaches by a large margin, obtaining a 129\% and 62\% relative improvement compared to NLLB-200 and GPT-4, respectively.
\end{abstract}

\section{Introduction}
The emergence of multilingual large language models (LLMs) and the wide availability of massive multilingual datasets have significantly advanced the field of Machine Translation (MT).
These developments have led to MT systems that not only perform exceptionally well in high-resource languages but also support a growing number of low-resource languages~\citep[\textit{inter alia}]{fan-etal-2021-beyond,tang-etal-2021-multilingual,costa-jussa-etal-2022-no, kudugunta-etal-2023-madlad}.
Nevertheless, the research community still faces several unresolved challenges in MT.\@
Among these, the translation of text that contains entities is still a hard task, especially with some categories of entities, e.g., movies, books, food, locations, and sometimes even people, to name a few. Indeed, \textit{word-for-word}, or \textit{literal}, translations of their names may not be suitable due to cultural-specific references, which can vary depending on social, geographical, historical, and political contexts, among other factors~\cite{hershcovich-etal-2022-challenges}.
Therefore, the challenge lies in accurately identifying when and how to translate entities whose names are significantly different across languages.
This step is crucial, as relying on literal translations may not convey the intended meaning, risking the effectiveness of the entire translation process~\cite{gaballo-2012-exploring,diaz-millon-olvera-lobo-2023-towards}.
For example, if we were to translate word-for-word ``Qual è la trama de \textit{Il Giovane Holden}?'' from Italian to English, we would obtain ``What is the plot of \textit{The Young Holden}?'', which is grammatically correct but semantically incorrect. The correct translation ``What is the plot of \textit{The Catcher in the Rye}?'' requires not only fluency in both the source and target languages but also knowledge of the cultural contexts involved.

In this paper, we address the problem of cross-cultural translation on two fronts: resources and methods. More specifically, our contributions can be summarized as follows:
\begin{itemize}
	\item We introduce \benchmark{}, the first large-scale, manually-created benchmark for cross-cultural translation across 10 language pairs of text containing entity names;
	\item We demonstrate that \benchmark{} exposes the limitations of current MT models and LLMs in translating text with entity names that can vary across languages and cultures;
	\item We propose KG-MT, a novel MT system equipped with retrieval-augmented generation from multilingual knowledge graphs;
	\item We evaluate KG-MT on \benchmark{} and show that it outperforms state-of-the-art approaches by a large margin, while also requiring minimal supervision and computational resources compared to data augmentation approaches.
\end{itemize}

\noindent We hope our work will encourage further research in the field of cross-cultural translation, leading to more investigations on the gaps of current methods in capturing cultural nuances beyond differences in entity names.

\section{Related Work}
MT is a long-standing research topic in NLP.\@
In this section, we briefly review the literature on recent advancements in MT, with a focus on studies that investigate entity names in relation to MT.

\paragraph{Machine Translation.} The field of MT has made a significant step forward with the emergence of multilingual language models, such as mBERT~\cite{devlin-etal-2019-bert} and XLM-R~\cite{conneau-etal-2020-unsupervised}, and massive multilingual corpora, such as OSCAR~\cite{ortiz-suarez-etal-2019-asynchronous} and MADLAD-400~\cite{kudugunta-etal-2023-madlad}.
Not only have these developments led to robust bilingual MT systems, such as OPUS-MT~\cite{tiedemann-etal-2023-democratizing}, but also to multilingual MT systems that can translate to and from multiple languages with a single model, such as mBART-50~\cite{liu-etal-2020-multilingual-denoising}, M2M-100~\cite{fan-etal-2021-beyond}, and NLLB-200~\cite{costa-jussa-etal-2022-no}.
Therefore, we build KG-MT on top of these multilingual MT systems -- which are openly available and widely used in the research community -- while also comparing our results with state-of-the-art LLMs, such as GPT-3.5 and GPT-4, which have been shown to achieve competitive performance in general-purpose MT evaluations~\cite{wang-etal-2023-document-level}.

\paragraph{External Knowledge in Machine Translation.}
Previous studies have already introduced methods to improve MT system via retrieval-augmentation or constrained-generation~\cite{zhang-etal-2018-guiding,bulte-tezcan-2019-neural,campolungo-etal-2022-reducing,iyer-etal-2023-code}.
Notably, \citet{zhang-etal-2018-guiding} proposed retrieval-augmentation to improve the translation of low-frequency words at inference time, while \citet{bulte-tezcan-2019-neural} demonstrated the benefits of retrieving fuzzy matches to augment a dataset at training time.
More recently, \citet{campolungo-etal-2022-reducing} and \citet{iyer-etal-2023-code} investigated the use of lexical constraints derived from external knowledge sources, e.g., dictionaries like WordNet, to improve the translation of senses in the long tail of the distribution.
Although guiding or constraining the translation process has been shown to be an effective direction towards improving the translation quality of MT systems, the area at the intersection of retrieval-augmented generation and retrieval from large knowledge sources with millions of elements, such as Wikidata, is still understudied, to the best of our knowledge.

\begin{figure*}[ht]
	\centering
	\includegraphics[width=\linewidth]{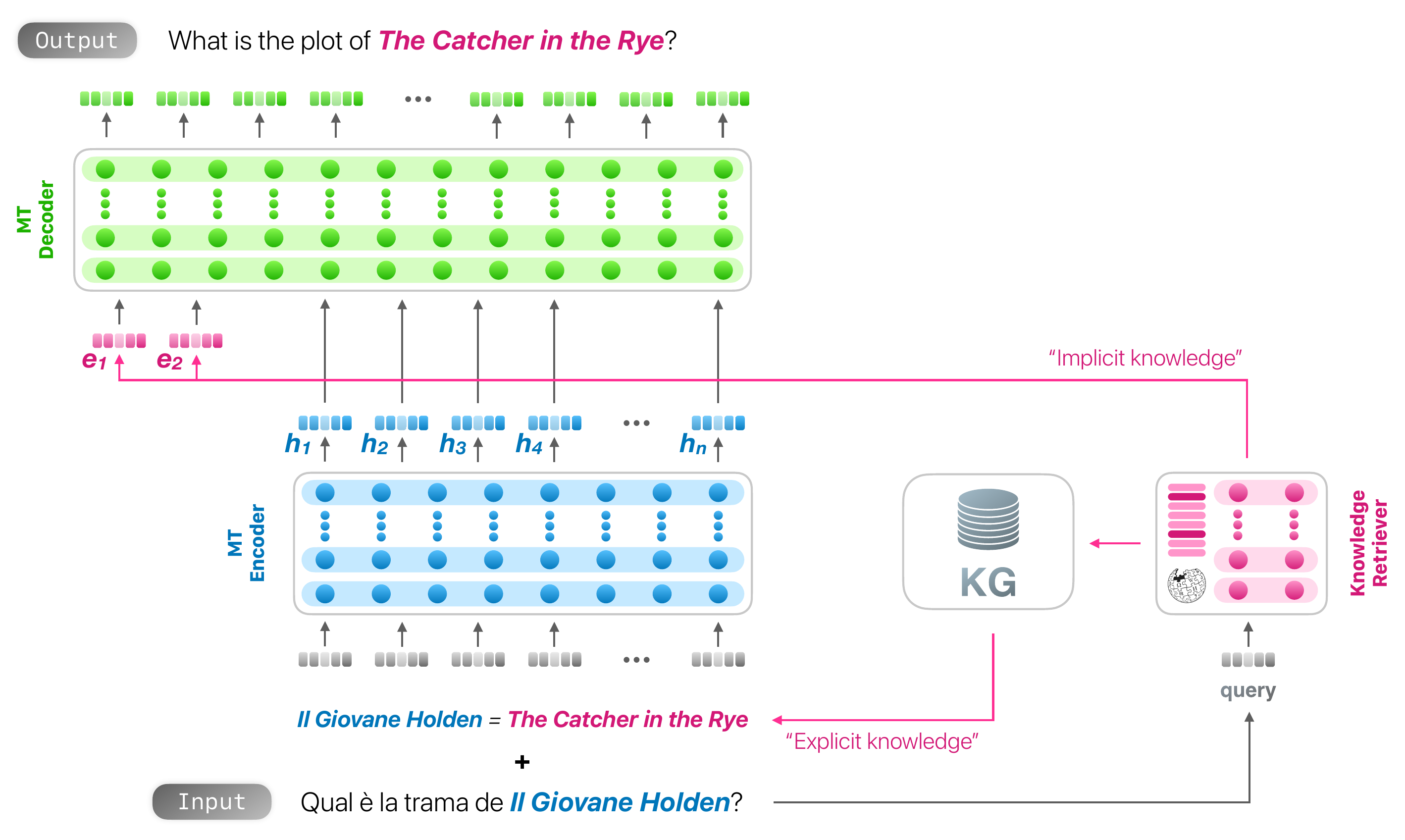}
	\caption{Overview of KG-MT, which leverages a \textit{knowledge retriever}, i.e., a dense retrieval mechanism to retrieve the most relevant entities from a multilingual knowledge graph (see Section~\ref{subsec:retriever}), to improve the translation.
		The retrieved entities are then integrated into the MT system in two ways: \textit{explicit} knowledge integration, where the entity names are explicitly added to the source text (see Section~\ref{subsec:generator}), and \textit{implicit} knowledge integration, where the entity embeddings are fused with the encoder hidden states (see Section~\ref{subsec:fusion}).}
	\label{fig:kgmt}
\end{figure*}

\paragraph{Entity names in Machine Translation.} Earlier investigations have long recognized and begun to address the challenges associated with translating texts that contain entity names~\cite{knight-graehl-1998-machine,al-onaizan-knight-2002-named,al-onaizan-knight-2002-translating}.
However, there are three important aspects that have yet to be fully explored in the literature.
First, the focus has predominantly been on the \textit{transliteration} of entity names, i.e., adapting an entity name from the script of one language to another~\cite{sadamitsu-etal-2016-name,ugawa-etal-2018-neural,zeng-etal-2023-extract}.
Although transliteration is crucial for languages with different scripts, like English and Chinese, it does not necessarily account for the \textit{transcreation} of entity names between languages using the same script,
like English and Italian~\cite{gaballo-2012-exploring,diaz-millon-olvera-lobo-2023-towards}.
Second, the depth of existing investigations has been constrained significantly by the absence of large-scale and high-quality benchmarks designed to highlight the challenges of cross-cultural translation~\cite{zeng-etal-2023-extract}.
Lastly, current approaches have mainly relied on training MT models by synthetically augmenting the training datasets to cover more entity names~\cite{liu-etal-2021-mulda,hu-etal-2022-deep,saleva-lignos-2022-paranames}.
However, data augmentation strategies, despite their effectiveness, often lead to a substantial increase of the training dataset size and the computational resources needed for training, especially when the entities to cover are in the millions.
Furthermore, they also require frequent retraining to incorporate names of emerging entities (e.g., new movies, books, shops, restaurants, and products) and update entities whose name has changed (e.g., new aliases, nicknames, and stage names).

Our work addresses the foregoing three aspects by introducing \benchmark{}, the first large-scale, manually-created benchmark for cross-cultural translation of text containing entity names, and by proposing KG-MT, a novel MT system that automatically retrieves the most relevant entity names from a multilingual knowledge graph and integrates them into the translation process on the fly, instead of memorizing every possible entity name translation for each source-target language pair through synthetic data augmentation.

\section{Enhancing Machine Translation using Multilingual Knowledge Graphs}

In contrast with augmentation strategies based on synthetic data that aim to maximize entity coverage at model training time, our hypothesis is that MT systems do not need to memorize every possible entity name transliteration and transcreation for each source-target language pair to correctly translate a text that contains entities.
Instead, the core idea that motivates our work is to leverage an external knowledge source to first retrieve the most relevant entities for an input text, and then generate the translation by incorporating the retrieved entity names in the target language.
In fact, multilingual knowledge graphs, such as DBPedia~\cite{auer-etal-2007-dbpedia}, BabelNet~\cite{navigli-ponzetto-2012-babelnet,navigli-etal-2021-ten}, and Wikidata~\cite{vrandecic-krotzsch-2014-wikidata}, provide a wealth of lexical and factual knowledge about millions of entities in many languages, including their names, aliases, and descriptions~\cite{kaffee-et-al-2023-multilingual,conia-etal-2023-increasing}.
Not only that, but such knowledge is also easier to edit and is frequently updated to reflect the latest changes in the real world. By leveraging a multilingual knowledge graph, the focus of our approach shifts from memorizing entity names to learning when and how to retrieve the most relevant entities for a given input text and integrate their names in the target language in an end-to-end fashion.

Our method, KG-MT, features two main components: (i) a \textit{knowledge retriever}, which retrieves the most relevant entities about the source text from a knowledge graph (Section~\ref{subsec:retriever}), and (ii) a \textit{knowledge-enhanced translator} that generates the target text by incorporating the retrieved entity names (Section~\ref{subsec:generator}).
To better model the interactions between the retrieved entities and the translation, we also introduce a method to fuse the representations of the two components (Section~\ref{subsec:fusion}).
Figure~\ref{fig:kgmt} provides an overview of KG-MT, which we describe in detail in the following sections.

\subsection{Retrieving Relevant Entities from Multilingual Knowledge Graphs}\label{subsec:retriever}
Given a source text $\mathbf{t} = \langle w_1, \dots, w_n \rangle$ in a source language $l_s$, the objective of our \textit{knowledge retriever} is to retrieve the top-$k$ most relevant entities $\mathcal{E}_\mathbf{t} = \{e_1, \dots, e_k\}$ from a knowledge graph $G =\mathcal{E} \times \mathcal{R} \times \mathcal{E}$, where $\mathcal{E}$ is the set of entities and $\mathcal{R}$ is the set of relations in the multilingual knowledge graph.\footnote{We use Wikidata as our reference multilingual knowledge graph in this work; however, our method is not limited to a specific knowledge graph and can be extended to other knowledge graphs with similar structure, such as BabelNet.}
We represent each entity $e_i$ as a tuple $e_i = \langle n_i, d_i \rangle$, where $n_i$ is the primary name of the entity and $d_i$ is its description.
Including the description of an entity allows us to distinguish between homonyms, i.e., entities with the same name.

We define the relevance score $s(e_i, \mathbf{t})$ of an entity $e_i$ with respect to the source text $\mathbf{t}$ as the cosine similarity between the entity and the source text:
\begin{equation}
	s(e_i, \mathbf{t}) = \frac{\mathbf{e}_i \cdot \mathbf{t}}{\|\mathbf{e}_i\| \|\mathbf{t}\|},
\end{equation}
where $\mathbf{e}_i$ is the embedding of the entity $e_i$.\@ We then retrieve the top-$k$ most relevant entities for the source text $\mathbf{t}$ as follows:
\begin{equation}
	\mathcal{E}_\mathbf{t} = \operatorname{top}_k\left(\{e_i \in \mathcal{E} \mid s(e_i, \mathbf{t})\}\right).
\end{equation}

The embedding $\mathbf{e}_i$ of an entity $e_i$ is obtained from an encoder~\cite{izacard-etal-2021-unsupervised}, which we train contrastively to maximize the likelihood of retrieving a relevant entity $e^+$ and minimize the likelihood of retrieving an irrelevant entity $e^-$ given $\mathbf{t}$ as the input query:
\begin{equation}
	\mathcal{L} = -\log \frac{\exp(\mathbf{e}^+ \cdot \mathbf{t})}{\exp(\mathbf{e}^+ \cdot \mathbf{t}) + \sum_{i=1}^{n} \exp(\mathbf{e}^-_i \cdot \mathbf{t})}
\end{equation}

\noindent where $\mathbf{e}^+$ is the embedding of a relevant entity $e^+$ and $\mathbf{e}^-$ is the embedding of an irrelevant entity $e^-$.
Importantly, we also introduce a sampling strategy to mine hard negative examples, i.e., instead of randomly sampling in-batch negatives~\cite{botha-etal-2020-entity}, we select $n$ homonymous entities that have the same name as the relevant entity $e^+$ but are not relevant to the source text $\mathbf{t}$.

\subsection{Integrating Explicit Knowledge into a Machine Translation Model}\label{subsec:generator}
Given the source text $\mathbf{t} = \langle w_1, \dots, w_n \rangle$ in a source language $l_s$ and the entities $\mathcal{E}_\mathbf{t}$ retrieved by our knowledge retriever, the objective of our \textit{knowledge-enhanced translator} is to generate the target text $\mathbf{t}' = \langle w'_1, \dots, w'_m \rangle$ in a target language $l_t$ by incorporating the entity names of $\mathcal{E}_\mathbf{t}$ into the translation process.
Therefore, instead of directly generating the target text $\mathbf{t}'$ from the source text $\mathbf{t}$, we first build a knowledge-enhanced source text $\mathbf{t}^{\textsc{+kg}}$ as follows:
\begin{align}
	\mathbf{t}^{\textsc{+kg}} = \langle & w_1, \dots, w_n, \nonumber                                                      \\
	                                    & [\textsc{KG}], n_1^s \rightarrow n_1^t, \dots, n_k^s \rightarrow n_k^t \rangle,
\end{align}

\noindent where $[\textsc{KG}]$ is a special token that indicates the start of the entity name translations, $n_i^s$ is the name of the entity $e_i$ in the source language $l_s$ and $n_i^t$ is the name of the entity $e_i$ in the target language $l_t$, as provided by the multilingual knowledge graph.
We then feed the knowledge-enhanced source text $\mathbf{t}^{\textsc{+kg}}$ to a standard sequence-to-sequence MT model to generate the target text $\mathbf{t}'$, in a similar vein to past work on guiding MT systems~\cite{zhang-etal-2018-guiding,bulte-tezcan-2019-neural}.
Given the format of $\mathbf{t}^{\textsc{+kg}}$, the MT model is fine-tuned to learn how to generate the target text $\mathbf{t}'$ by also attending to the translation of the entity names $n_i^t$.
We refer to this method as \textit{explicit} knowledge integration, as the translations of the relevant entities are explicitly provided in the input to the MT model.

\subsection{Integrating Implicit Knowledge into a Machine Translation Model}\label{subsec:fusion}
Although the knowledge-enhanced translator can generate the target text
$\mathbf{t}'$ by incorporating the entity names of $\mathcal{E}_\mathbf{t}$, it does not take advantage of the representations of the retrieved entities $\mathbf{e}_i$ learned by the knowledge retriever.
To overcome this limitation, we also propose a method to fuse the latent representations of the knowledge retriever and the knowledge-enhanced translator, which allows KG-MT to better model the interconnections between the retrieved entities and the generated translation.

Here, we assume that an MT model is structured as an encoder-decoder architecture, such as the Transformer model~\cite{vaswani-etal-2017-attention}.
Given a general encoder-decoder architecture, we feed the knowledge-enhanced source text $\mathbf{t}^{\textsc{+kg}}$ to the encoder of the MT model to obtain the encoder hidden states $\mathbf{h}^{\textsc{+kg}} = \langle \mathbf{h}_1, \dots, \mathbf{h}_{n+k+1} \rangle$, where $\mathbf{h}_i$ is the hidden state of the encoder at position $i$.\footnote{To simplify the notation, we assume that each entity name translation $n_i^s \rightarrow n_i^t$ is represented by a single token, even though it is actually composed of at least three tokens.}
Then, we prepend the embeddings $\mathbf{e}_i$ of the retrieved entities to the encoder hidden states $\mathbf{h}^{\textsc{+kg}}$ to obtain the encoder hidden states $\mathbf{h}^{\textsc{+kg+e}}$:
\begin{equation}
	\mathbf{h}^{\textsc{+kg+e}}
	= \langle \mathbf{e}_1, \dots, \mathbf{e}_k, \mathbf{h}_1, \dots, \mathbf{h}_{n+k+1} \rangle.
\end{equation}

\noindent Finally, we feed the hidden states $\mathbf{h}^{\textsc{+kg+e}}$ to the decoder of the MT model, which is now able to also attend to the entity embeddings from the retriever and fuse them with its hidden states.
Our intuition is that the embeddings $\mathbf{e}_i$ from the knowledge retriever can contain useful fine-grained, latent information about the retrieved entities.
We refer to this method as \textit{implicit} knowledge integration, using an embedding-based fusion strategy reminiscent of Fusion-in-Decoder~\cite[FiD]{izacard-grave-2021-distilling,izacard-grave-2021-leveraging}.
However, unlike FiD, our knowledge-enhanced translator fuses the hidden states of two different encoders, i.e., the knowledge retriever and the encoder of the knowledge-enhanced translator, as shown in Figure~\ref{fig:kgmt}.


\section{Evaluating Cross-Cultural Translation of Texts Containing Entity Names}\label{sec:benchmark}
To evaluate the effectiveness of our method, we introduce Cross-Culture Translate (\benchmark{}), the first large-scale, manually-curated benchmark for the task of cross-cultural translation of texts containing entity names.
\benchmark{} is composed of parallel texts in 10 English-to-X language pairs from a diverse set of languages, including both high-resource and low-resource languages, namely, Arabic, Chinese, French, German, Italian, Japanese, Korean, Spanish, Thai, and Turkish.
We highlight that this design choice allows our benchmark to feature languages with diverse scripts, some of which are similar to English, such as French and Spanish, and others that are very different, such as Arabic, Chinese, and Thai.
Importantly, our benchmark is:
\begin{itemize}
	\item \textbf{Challenging:} \benchmark{} is the first benchmark to focus on cross-cultural translation of texts containing entity names, which is particularly challenging due to the cultural-specific references of entity names across languages;
	\item \textbf{Large-scale:} \benchmark{} contains about 5,000 sentences for each language pair for a total of over 58,000 instances, making it one of the largest benchmarks for MT, independently of its focus on cross-cultural translation;
	\item \textbf{Multi-reference:} \benchmark{} provides multiple translations for each source text (over 100,000 references, or 2 translations per sentence on average);
	\item \textbf{Gold-quality:} \benchmark{} is manually created and verified by human annotators fluent in the source and target languages, which ensures the quality of the benchmark and the correctness of the translations.
\end{itemize}
\noindent We believe that \benchmark{} will be a valuable resource for the MT research community and will encourage further research on the problem of cross-cultural translation of texts containing entity names.

\subsection{Design Principles}\label{subsec:benchmark-principles}
The creation process of \benchmark{} is mainly driven by two design principles: (i) the texts should contain entity names that are likely to be affected not only by transliteration between languages, and (ii) the heart of the challenge should be the translation of the entity names, rather than the translation of the rest of the text, i.e., the text should not be too complex to translate if the entity names are translated correctly.

To satisfy the first design principle, we first identify for each language pair a set of entities from Wikidata that adhere to the following two main criteria: (a) the entity has at least one name in English and one name in the target language, (b) the English name of the entity is at least 50\% different from the names in French, German, Italian, Spanish, and their word-for-word translation to English, as measured by the Levenshtein distance.
The rationale behind the second criterion is that such a difference in the entity names across languages that mostly share the same script is likely an indicator of a name dissimilarity that goes beyond transliteration.
For example, the Italian name of the entity ``\textit{The Catcher in the Rye}'' is ``\textit{Il Giovane Holden}'', while its French name is ``\textit{L'Attrape-c\oe{}urs}''.

To satisfy the second design principle, we ask a group of human annotators to curate a set of short knowledge-seeking questions -- less than 25 words -- in English about the identified entities. Requiring the question to be short encourages simple and concise questions that are easy to translate if the entity names are correctly translated.
Moreover, requiring the text to be a question mitigates the risk of including inaccurate facts in the text: for example, ``Is \textit{The Catcher in the Rye} a book by J. D. Salinger?'' is a legitimate question, while ``\textit{The Catcher in the Rye} is a book by J. D. Salinger'' is a factual statement that may or may not be factually accurate.

\subsection{Translation Process}
Having identified the entities of interest for each language pair and having created English questions about them, we produce the translations in each target language via a two-step process.
First, we ask a group of human translators to translate the questions from English to the target language.
Then, we ask a second group of human annotators to verify the correctness of the translations.\footnote{The intersection between the translators and the annotators is not empty but is kept to a minimum to avoid bias, and the probability of an annotator verifying their own translation is low.}

The entire process is guided by a set of instructions and guidelines that we provide to the annotators. Moreover, we require the annotators to be fluent in English, native speakers of the target language, and resident in a country where the target language is spoken.
Before starting the translation process, we also require the annotators to pass an entrance test to further verify their language proficiency and their comprehension of the instructions and guidelines; otherwise, they are not allowed to participate in the annotation task.
Finally, the annotators are periodically evaluated on a set of test questions: if they fail on them, they are excluded from the pool of annotators.
Since each English question is formulated from a given entity, we can aid the translators by providing the entity name(s) from Wikidata in the target language as a hint (see Design Principle \textit{i.a} in Section~\ref{subsec:benchmark-principles}), the English and target language descriptions of the entity from Wikidata, and the English and target language Wikipedia pages of the entity, which are fundamental resources to understand the context and background of the entity of interest.

At the end of the process, each English question is translated into the target language by at least three different translators, and each translation is then verified by at least three different annotators, allowing us to retain only the translations that are agreed upon by the annotators.
We provide more details about this process in the Appendix.

\subsection{Evaluation Metrics}
It is well known that the evaluation of MT systems is challenging, as there is no single metric that can capture all the aspects of translation quality.
For example, BLEU~\cite{papineni-etal-2002-bleu} is a popular metric that measures the n-gram overlap between the generated translation and the reference translations, but it is long known not to correlate strongly with human judgments~\cite{callison-burch-etal-2006-evaluating}.
More recently, the research community has proposed alternative metrics, such as BERTScore~\cite{zhang-etal-2020-bertscore} and
COMET~\cite{rei-etal-2020-comet}, that aim to capture more nuanced aspects of translation quality, such as semantic similarity and factual correctness.
However, such learned metrics yield only a translation-level score, which is not easy to interpret~\cite{perrella-etal-2024-beyond} and does not allow us to easily analyze the translation at the entity level.

To address the foregoing limitations, not only do we provide the translations of the questions in \benchmark{}, but also the list of the valid translations of the entity names that are valid in the context of the considered text.
Having a comprehensive list of manually-curated valid names allows us to introduce M-ETA (Manual Entity Translation Accuracy), a simple metric to easily measure the translation quality at the entity level.
Differently from previuos metrics that rely on automatically identifying and aligning entities~\cite{hu-etal-2022-deep}, M-ETA directly checks whether an automatic translation contains one of the manually-curated names.
More formally, given a translation $\mathbf{t}'$ in a target language $l_t$ and a set of gold entities $\hat{\mathcal{E}}_\mathbf{t'}$, we define the entity-level translation quality score $Q(\mathbf{t}', \hat{\mathcal{E}}_\mathbf{t'})$ as follows:
\begin{equation}
	Q(\mathbf{t}', \hat{\mathcal{E}}_\mathbf{t'}) = \frac{1}{|\hat{\mathcal{E}}_\mathbf{t'}|} \sum_{e_i \in \hat{\mathcal{E}}_\mathbf{t'}} q(\mathbf{t}', e_i),
\end{equation}
where $q(\mathbf{t}', e_i)$ is the entity-level translation quality score of the entity $e_i$ in the target text $\mathbf{t}'$ and is defined as follows:
\begin{equation}
	q(\mathbf{t}', e_i) = \min \{1, \sum_{n_i^t \in \mathcal{N}_{e_i}^t} \mathbb{I}(n_i^t \in \mathbf{t}') \},
\end{equation}
where $\mathcal{N}_{e_i}^t$ is the set of manually-curated names of the entity $e_i$ in the target language $l_t$ and $\mathbb{I}(n_i^t \in \mathbf{t}')$ is an indicator function that is equal to 1 if the name $n_i^t$ of the entity $e_i$ is in the target text $\mathbf{t}'$ and 0 otherwise.

\section{Experiments and Results}\label{sec:experiments}
In this section, we first list the systems we consider in our main experiments, then describe the datasets used to train the MT systems, and finally report and discuss the results.

\paragraph{Systems.} We compare the following systems:
\begin{itemize}
	\item \textbf{GPT-3}, \textbf{GPT-3.5}, and \textbf{GPT-4}:\footnote{\ Timestamps: GPT-3 -- \texttt{text-davinci-003}, GPT-3.5 -- \texttt{gpt-3.5-turbo-0613}, GPT-4 -- \texttt{gpt-4-0613}.} among the most popular and best performing LLMs, which have shown strong translation performance in the literature;
	\item \textbf{mBART-50}, \textbf{M2M-100}, and \textbf{NLLB-200}: recent multilingual MT models that support translation from and to about 50, 100, and 200 different languages using a single model, respectively;
	\item \textbf{KG-MT}: our proposed approach, which leverages the information available in multilingual knowledge graphs for end-to-end retrieval-augmented translation.
\end{itemize}

\noindent To ensure a fair comparison, we fine-tune mBART-50, M2M-100, NLLB-200, and KG-MT using the same dataset.
For more details about our experimental setup, please refer to the Appendix.

\paragraph{Datasets.} As mentioned in Sections~\ref{subsec:retriever} and~\ref{subsec:generator}, KG-MT requires two datasets for training: one for the knowledge retriever and one for the knowledge-enhanced translator.
The training dataset for retrieval should contain instances of the form $\langle \mathbf{t}, e^+, e^- \rangle$, where $\mathbf{t}$ is a source text, $e^+$ is a relevant entity for $\mathbf{t}$, and $e^-$ is an irrelevant entity for $\mathbf{t}$.
The training dataset for translation should contain instances of the form $\langle \mathbf{t}, \mathbf{t}', \hat{\mathcal{E}}_\mathbf{t} \rangle$, where $\mathbf{t}$ is a source text, $\mathbf{t}'$ is a target text, and $\hat{\mathcal{E}}_\mathbf{t}$ is the set of gold entities for $\mathbf{t}$.
To train the knowledge retriever, we use the training data from Mintaka~\cite{sen-etal-2022-mintaka}, a recently proposed dataset for multilingual question answering in which each question is tagged with the entities that appear therein.
Since the questions in Mintaka are manually translated, we can also use its gold translations to train the knowledge-enhanced translator.

\begin{table}[t]
	\centering
	\resizebox{\linewidth}{!}{%
		\begin{tabular}{clcccc}
			\toprule
			                                              &                        & \multicolumn{3}{c}{EN-to-XX (Avg)} &                                                                      \\
			\cmidrule(lr){3-5}
			                                              &                        & \textsc{\small BLEU}               & \textsc{\small COMET} & \textsc{\small M-ETA} & Size                 \\
			\midrule
			\multirow{3}{*}{\rotatebox{90}{\textit{LLM}}} & GPT-3                  & 37.4                               & 75.4                  & 14.1                  & \textsc{\small 175B} \\
			                                              & GPT-3.5                & 42.8                               & 77.8                  & 20.9                  & \textsc{\small ?}    \\
			                                              & GPT-4                  & 50.9                               & 82.1                  & 25.3                  & \textsc{\small ?}    \\
			\midrule
			\multirow{3}{*}{\rotatebox{90}{\textit{MT}}}  & mBART-50               & 36.1                               & 79.8                  & 12.2                  & \textsc{\small 0.6B} \\
			                                              & M2M-100                & 34.8                               & 77.9                  & 11.5                  & \textsc{\small 0.4B} \\
			                                              & NLLB-200               & 39.5                               & 81.9                  & 17.9                  & \textsc{\small 0.6B} \\
			\midrule
			\multirow{3}{*}{\rotatebox{90}{\textit{RAG}}} & KG-MT$_\textrm{mBART}$ & 44.1                               & 79.7                  & 39.1                  & \textsc{\small 0.6B} \\
			                                              & KG-MT$_\textrm{M2M}$   & 42.6                               & 80.8                  & 38.3                  & \textsc{\small 0.4B} \\
			                                              & KG-MT$_\textrm{NLLB}$  & \textbf{51.8}                      & \textbf{84.6}         & \textbf{41.1}         & \textsc{\small 0.6B} \\
			\bottomrule
		\end{tabular}
	}%
	\caption{Average performance of the baselines and the variants of KG-MT on \benchmark{} compared to the state-of-the-art multilingual MT systems and LLMs.}\label{tab:results-average}
\end{table}

\begin{table*}[t]
	\centering
	\resizebox{\textwidth}{!}{%
		\begin{tabular}{l*{20}{c}}
			\toprule
			                       & \multicolumn{2}{c}{EN-AR} & \multicolumn{2}{c}{EN-DE} & \multicolumn{2}{c}{EN-ES} & \multicolumn{2}{c}{EN-FR} & \multicolumn{2}{c}{EN-IT} & \multicolumn{2}{c}{EN-JA} & \multicolumn{2}{c}{EN-KO} & \multicolumn{2}{c}{EN-TH} & \multicolumn{2}{c}{EN-TR} & \multicolumn{2}{c}{EN-ZH}                                                                                                                                                                                                                                            \\
			\cmidrule(lr){2-3} \cmidrule(lr){4-5} \cmidrule(lr){6-7} \cmidrule(lr){8-9} \cmidrule(lr){10-11} \cmidrule(lr){12-13} \cmidrule(lr){14-15} \cmidrule(lr){16-17} \cmidrule(lr){18-19} \cmidrule(lr){20-21}
			                       & \textsc{\small bleu}      & \textsc{\small m-eta}     & \textsc{\small bleu}      & \textsc{\small m-eta}     & \textsc{\small bleu}      & \textsc{\small m-eta}     & \textsc{\small bleu}      & \textsc{\small m-eta}     & \textsc{\small bleu}      & \textsc{\small m-eta}     & \textsc{\small bleu} & \textsc{\small m-eta} & \textsc{\small bleu} & \textsc{\small m-eta} & \textsc{\small bleu} & \textsc{\small m-eta} & \textsc{\small bleu} & \textsc{\small m-eta} & \textsc{\small bleu} & \textsc{\small m-eta} \\
			\midrule
			mBART-50               & 21.7                      & 10.9                      & 45.5                      & 18.5                      & 44.9                      & 15.1                      & 44.3                      & 15.6                      & 43.2                      & 12.9                      & 43.9                 & 13.9                  & 26.7                 & 8.3                   & 45.6                 & 1.9                   & 26.2                 & 20.5                  & 18.5                 & ~~4.1                 \\
			M2M-100                & 27.3                      & 13.5                      & 37.9                      & 11.5                      & 51.7                      & 19.7                      & 41.8                      & 13.7                      & 44.9                      & 17.4                      & 34.6                 & ~~8.0                 & 28.1                 & ~~8.0                 & 35.8                 & 0.8                   & 29.6                 & 19.3                  & 16.0                 & ~~2.8                 \\
			NLLB-200               & 25.2                      & 20.5                      & 43.1                      & 19.6                      & 63.6                      & 31.5                      & 52.3                      & 24.7                      & 55.4                      & 26.4                      & 29.0                 & ~~8.4                 & 32.6                 & 17.7                  & 42.6                 & 1.8                   & 34.3                 & 25.4                  & 17.2                 & ~~3.1                 \\
			\midrule
			GPT-3                  & 18.9                      & 15.1                      & 42.7                      & 18.1                      & 55.0                      & 26.2                      & 49.1                      & 22.8                      & 53.1                      & 22.0                      & 36.9                 & ~~9.6                 & 26.3                 & 8.2                   & 37.8                 & 2.2                   & 20.5                 & 11.7                  & 33.4                 & ~~5.1                 \\
			GPT-3.5                & 22.7                      & 16.2                      & 46.3                      & 25.5                      & 62.2                      & 33.8                      & 56.0                      & 28.2                      & 55.4                      & 29.5                      & 41.0                 & 18.7                  & 31.2                 & 14.9                  & 41.1                 & 2.5                   & 29.8                 & 23.0                  & 42.5                 & 16.7                  \\
			GPT-4                  & 34.9                      & 23.8                      & 54.8                      & 27.8                      & 67.2                      & 35.2                      & 61.3                      & 28.5                      & 62.7                      & 34.7                      & 48.7                 & 22.9                  & 44.4                 & 18.4                  & 42.1                 & 5.4                   & 44.8                 & 36.4                  & 48.0                 & 19.9                  \\
			\midrule
			KG-MT$_\textrm{mBART}$ & 32.6                      & 48.5                      & 48.8                      & 41.5                      & 60.5                      & 47.7                      & 50.6                      & 41.8                      & 58.8                      & 51.3                      & 49.8                 & 48.5                  & 22.2                 & 36.9                  & 48.2                 & 12.7                  & 27.3                 & 48.1                  & 20.9                 & 13.5                  \\
			KG-MT$_\textrm{M2M}$   & 35.0                      & 45.3                      & 43.7                      & 39.4                      & 62.4                      & 49.1                      & 51.5                      & 41.3                      & 59.3                      & 51.4                      & 46.6                 & 40.7                  & 35.4                 & 29.9                  & 12.8                 & 12.7                  & 40.0                 & 62.2                  & 21.8                 & 11.0                  \\
			KG-MT$_\textrm{NLLB}$  & 37.3                      & 50.6                      & 55.8                      & 36.5                      & 68.9                      & 47.8                      & 58.7                      & 39.8                      & 65.3                      & 47.5                      & 48.6                 & 42.2                  & 43.8                 & 47.1                  & 56.7                 & 39.6                  & 41.2                 & 49.7                  & 20.9                 & 10.6                  \\
			\bottomrule
		\end{tabular}%
	}
	\caption{Performance by language pair of the baselines and the variants of KG-MT on \benchmark{} compared to the state-of-the-art multilingual MT systems and LLMs.}
	\label{tab:results}
\end{table*}


\paragraph{Results on \benchmark{}.} Table~\ref{tab:results-average} shows the results of the systems averaged over all the language pairs of \benchmark{} in terms of BLEU, COMET, and M-ETA.\@ We can first observe that MT systems, such as mBART-50, M2M-100, and NLLB-200, as well as LLMs, such as GPT-3, GPT-3.5, and GPT-4, obtain unsatisfactory M-ETAs on \benchmark{}, with NLLB-200 and GPT-4 achieving the highest average score of 17.9\% and 25.3\%, respectively.
These results support two of our hypotheses: (i) translating texts that contains challenging entity names is particularly difficult, and simply trying to translate the original entity name is often not sufficient to produce a correct translation; and (ii) BLEU and COMET are not a reliable metrics to evaluate the translation quality in this setting.
Indeed, a translation that is correct except for the entity names still receives high BLEU and COMET scores, even though the error in the translation of the entity name may completely alter the meaning and the intent of the entire translation.

Table~\ref{tab:results-average} also shows that KG-MT outperforms all the baselines by a large margin, with an average M-ETA score of 41.1\% for its best variant, which is equivalent to a 129.1\% and 62.5\% relative improvement over the best MT system (NLLB-200) and the best LLM (GPT-4), respectively.
Most notably, KG-MT is capable of closing the gap between NLLB-200 and GPT-4, outperforming an LLM that is supposedly 100 times larger in terms of number of parameters.

The jump in performance for KG-MT is consistent across different underlying MT models, i.e., we observe similar M-ETA scores (39.1\%, 38.3\%, and 41.1\%) for the KG-MT variants independently of whether we use mBART-50, M2M-100, or NLLB-200 as the backbone for the knowledge-enhanced translation model in KG-MT.\@ This trend empirically demonstrates that our method is able to retrieve relevant entities about the source text and integrate the information available in multilingual knowledge graphs to improve the output quality.
In general, we observe that the improvement in M-ETA is also consistent across different language pairs, as shown in Table~\ref{tab:results}.


\paragraph{Results on WMT benchmarks.}
Having evaluated the performance of KG-MT on \benchmark{}, we now turn our attention to WMT benchmarks to assess whether our method degrades the translation quality in general-purpose MT benchmarks. To this end, we evaluate the performance of KG-MT on the English-to-X test sets from WMT17, WMT18, WMT19, WMT20, and WMT21.
Table~\ref{tab:results-wmt} shows the results of KG-MT on the WMT benchmarks in terms of BLEU and COMET. As we can observe, KG-MT achieves competitive BLEU and COMET scores on the WMT benchmarks compared to the MT baselines, which suggests that our method does not degrade the quality of general-purpose translations.
On the contrary, KG-MT generally achieves slightly improved BLEU and COMET scores compared to vanilla MT systems, e.g., KG-MT$_\textrm{NLLB}$ obtains an absolute improvement of 0.8 points in BLEU and 1.7 points in COMET over NLLB-200.

\begin{table}[t]
	\centering
	\resizebox{0.75\linewidth}{!}{%
		\begin{tabular}{clcc}
			\toprule
			                                              &                        & \multicolumn{2}{c}{EN-to-XX (Avg)}                  \\
			\cmidrule(lr){3-4}
			                                              &                        & \textsc{Bleu}                      & \textsc{Comet} \\
			\midrule
			\multirow{3}{*}{\rotatebox{90}{\textit{LLM}}} & GPT-3                  & 18.1                               & 48.1           \\
			                                              & GPT-3.5                & 22.2                               & 57.3           \\
			                                              & GPT-4                  & \textbf{24.4}                      & 61.0           \\
			\midrule
			\multirow{3}{*}{\rotatebox{90}{\textit{MT}}}  & mBART-50               & 22.0                               & 55.7           \\
			                                              & M2M-100                & 21.4                               & 52.8           \\
			                                              & NLLB-200               & 23.3                               & 60.2           \\
			\midrule
			\multirow{3}{*}{\rotatebox{90}{\textit{RAG}}} & KG-MT$_\textrm{mBART}$ & 22.8                               & 55.2           \\
			                                              & KG-MT$_\textrm{M2M}$   & 22.3                               & 54.3           \\
			                                              & KG-MT$_\textrm{NLLB}$  & 24.1                               & \textbf{61.9}  \\
			\bottomrule
		\end{tabular}
	}%
	\caption{Evaluation results (BLEU and COMET) averaged over the EN-to-XX test sets of WMT-17, WMT-18, WMT-19, WMT-20, and WMT-21.}
	\label{tab:results-wmt}
\end{table}

\section{Analysis and Discussion}\label{sec:analysis}
In this section, we analyze KG-MT, and discuss where we believe it may be improved in future work. We expand on this analysis in the Appendix.

\begin{figure}[t]
	\centering
	\includegraphics[width=0.9\linewidth]{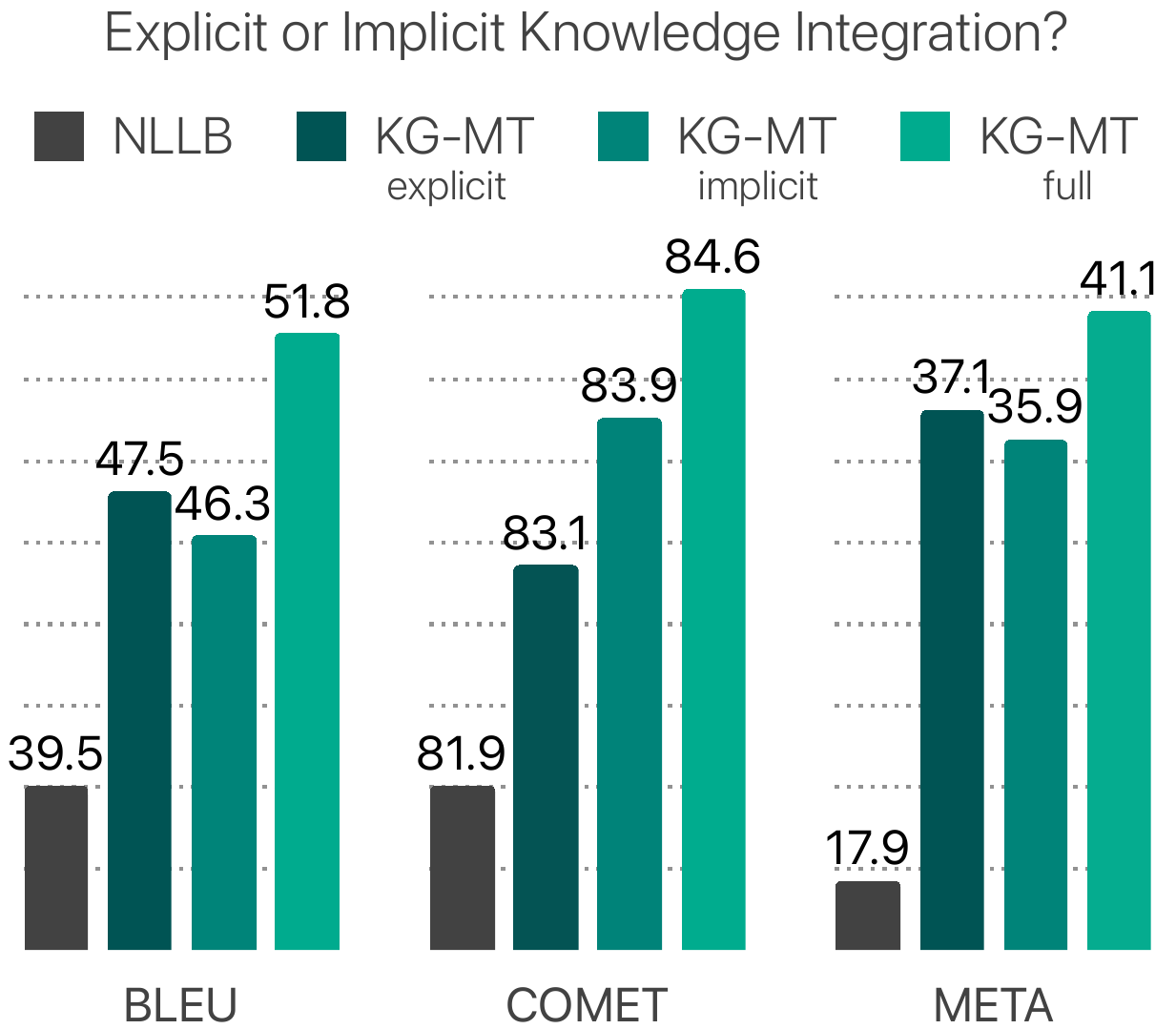}
	\caption{Results of KG-MT when using explicit or implicit knowledge integration, or both.}
	\label{fig:explicit-implicit-knowledge}
\end{figure}

\paragraph{Explicit or implicit knowledge?}
Sections~\ref{subsec:generator} and~\ref{subsec:fusion} introduce two methods to integrate the knowledge retrieved by the knowledge retriever into the knowledge-enhanced translator: explicit knowledge integration and implicit knowledge integration.
The first method is more straightforward but it also increases the length of the source text, which is undesirable since most of the attention mechanisms in popular Transformer-based models are quadratic with respect to the input length.
Instead, the second method requires intervening on the inner workings of the MT model, which is more complex and may not be always feasible.
However, it is also more flexible since the input length of the decoder only grows with the number of retrieved entities, which can be controlled by a hyperparameter.
To understand which method is more effective, we compare the performance of KG-MT when using explicit knowledge integration, implicit knowledge integration, and both.
Figure~\ref{fig:explicit-implicit-knowledge} shows that not only both methods are effective, but they are also complementary, as the combination of the two methods yields the best results.
We hypothesize that the injection of the entity embeddings with the implicit knowledge integration may also act as an indicator of whether the MT model should rely on its parametric memory for during the generation of the translation, independently of the semantics represented within the entity embeddings.

\paragraph{Knowledge retrieval.}
The knowledge retriever plays a fundamental role in KG-MT.\@ If the retrieval step collects wrong or unrelated entities, the knowledge-enhanced translator has to (i) be robust against noisy or irrelevant knowledge, and (ii) fall back on its parametric memory, which is often unreliable as shown by the results of the vanilla MT systems on \benchmark{} in Tables~\ref{tab:results-average} and~\ref{tab:results}. However, our analysis shows that our knowledge retriever achieves 85.9\% and 92.1\% hits@1 and hits@3, respectively, on \benchmark{}, which suggests that the knowledge retriever is effective at retrieving relevant entities. Part of this success can be attributed to our fine-tuning strategy with hard negative mining, which allows the knowledge retriever to improve the hits@1 and hits@3 by 5.6\% and 4.2\%, respectively, compared to using mContriever~\cite{izacard-etal-2021-unsupervised}, a pretrained retriever.
Given the good performance with hits@3, we set the knowledge retriever to retrieve at most three entities for each source text, which is a trade-off between retrieving more relevant entities and increasing the computational cost.


\begin{figure}[t]
	\centering
	\includegraphics[width=0.9\linewidth]{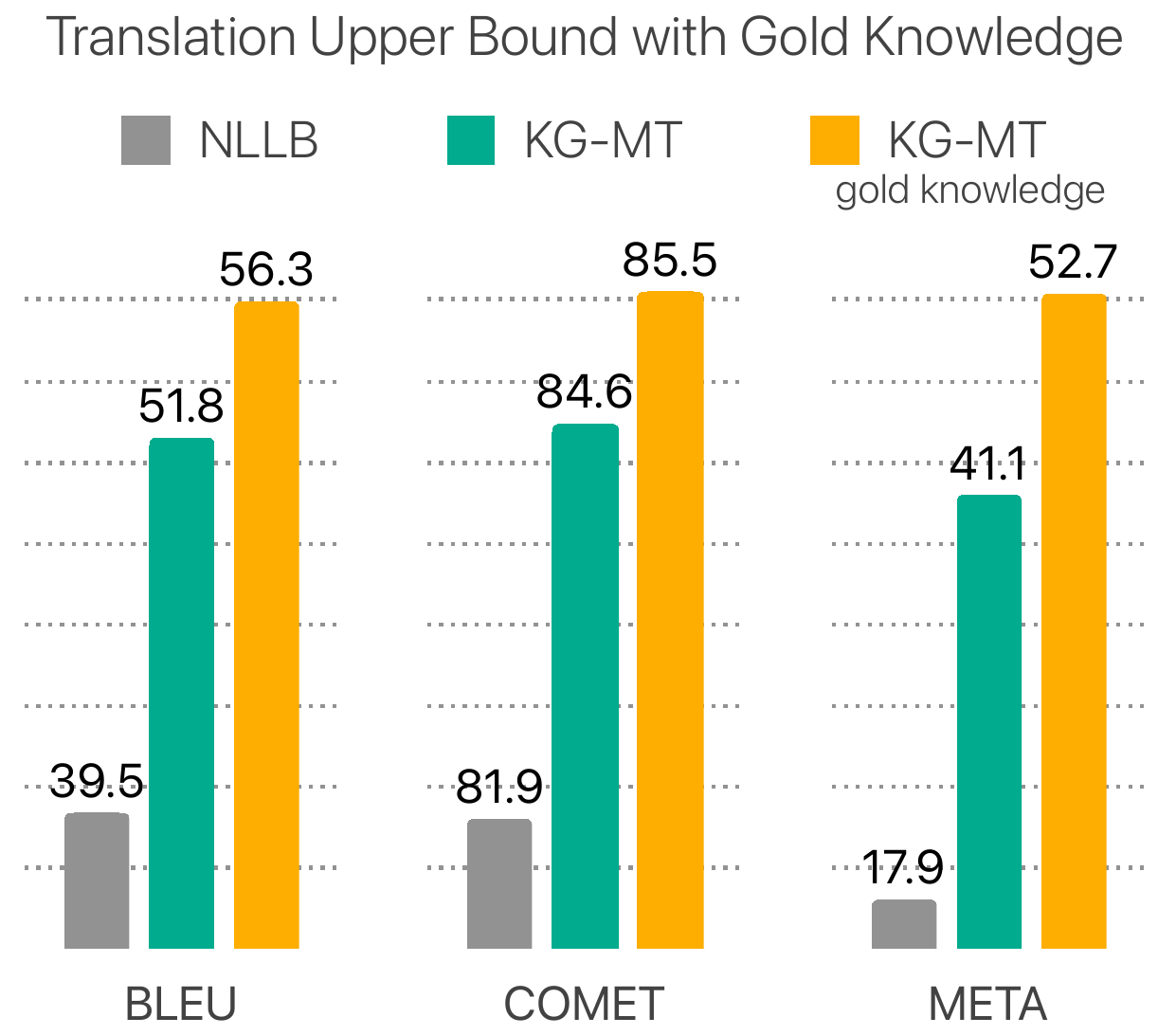}
	\caption{Results of KG-MT when using gold knowledge instead of the knowledge from the retriever.}
	\label{fig:knowledge-integration}
\end{figure}

\paragraph{Gold knowledge.}
If knowledge retrieval is not a significant source of errors for KG-MT, then the knowledge integration step, in which the knowledge-enhanced translator has to learn how to effectively integrate the retrieved entities into the translation process, is likely to be the main bottleneck.
To isolate the performance of the knowledge-enhanced translator from the performance of the knowledge retriever, we evaluate KG-MT when using gold entities instead of the entities retrieved by the knowledge retriever.
Figure~\ref{fig:knowledge-integration} shows that the performance of KG-MT increases only by 11.6\% in terms of average M-ETA when using gold entities, indicating that the knowledge-enhanced translator is not always capable of using the gold knowledge.\@ Therefore, we hypothesize that the primary area of gain for future work shall be on improving the knowledge integration step, e.g., by using more sophisticated fusion strategies for knowledge integration or creating better datasets for fine-tuning the MT model on the knowledge integration step.

\section{Conclusion and Future Work}\label{sec:conclusion}
In this paper, we addressed the problem of cross-cultural translation of texts containing entity names. Our contributions are threefold: (i) we introduced KG-MT, a novel approach for retrieval-augmented translation that leverages the information available in multilingual knowledge graphs to improve the translation of texts containing entity names; (ii) we introduced \benchmark{}, the first large-scale, manually-curated benchmark for the task of cross-cultural translation of texts containing entity names; and (iii) we conducted extensive experiments on \benchmark{} and other existing benchmarks for MT, showing that KG-MT significantly outperforms the state of the art on \benchmark{} while maintaining comparable results on general-purpose MT benchmarks.
We believe that our contribution will encourage further research on the problems that arise when translating texts containing cultural references beyond entity names, such as idioms and metaphors, and that retrieval-augmented translation will be a valuable tool to address these challenges.

\section*{Limitations}
In this section, we discuss some of the main limitations of our work and
how future research may be able to address them.

\paragraph{Language coverage.}
\benchmark{} contains a diverse set of languages, but it is clearly not exhaustive.
Although the number of languages included in our benchmark is comparable with the number of languages studied every year in the WMT shared tasks, it is still a small fraction of the world's languages.
While full coverage is likely infeasible, we still miss entire linguistic families, such as the Uralic, Dravidian, and Niger-Congo families, and many languages from the Indo-European family, such as Russian, Portuguese, and Hindi.
Future work should consider expanding the coverage of \benchmark{} to include more languages.
Indeed, different languages may present different challenges for cross-cultural translation, and it is important to understand these differences to develop more robust and generalizable translation systems.
Not only that, but another aspect that we do not consider in our work is the dialectal and regional variation within a language, which can also be a significant source of
errors in translation.
Our intuition is that, since current state-of-the-art MT systems and LLMs struggle in our setting, which mostly includes high- to medium-resource languages, they would struggle even more in low-resource languages and dialects.

\paragraph{Entity coverage and selection.}
The entities in \benchmark{} are selected from Wikidata, which is a large and diverse knowledge graph, but it is not complete. While our design principles (see Section~\ref{subsec:benchmark-principles}) are aimed at selecting entities that are likely to be challenging for MT systems, our selection strategy can also be considere aggressive for several reasons: (i) we only consider entities that are linked to Wikipedia pages, which may exclude many entities that are relevant in a given culture; (ii) we only consider entities that have at least an English name, which may exclude many entities that are relevant in a given culture; and (iii) our selection strategy is based on our experience with using Wikidata and Wikipedia.
Future work should consider more sophisticated strategies for selecting entities, such as considering language pairs that do not involve English and tuning the selection based on each language pair, rather than having a one-size-fits-all approach. Moreover, future work may also consider using other knowledge graphs, as Wikidata inherits the biases and errors of Wikipedia (and its editor demographics).

\paragraph{Translation quality.}
Our evaluation of KG-MT is mostly based on the M-ETA metric, which is a simple metric that measures the translation quality at the entity level.
While M-ETA is a useful metric to evaluate the performance of KG-MT, it is not a comprehensive metric to evaluate the translation quality of MT systems, i.e., it cannot be used alone to compare the performance different systems.
This is the reason why we also report the BLEU and COMET scores of KG-MT on \benchmark{} and the WMT benchmarks.
However, we acknowledge that BLEU and COMET are also not comprehensive metrics to evaluate the translation quality of MT systems.
Future work may consider fine-tuning learned metrics on our \benchmark{} annotations, which also include a list of manually-curated valid translations of the entity names that are valid in the context of the considered text.

\paragraph{Knowledge retrieval.}
Our knowledge retriever is based on a retrieval model that retrieves at most three entities for each source text. While this design choice is based on a trade-off between retrieving more relevant entities and increasing the computational cost, it is not clear whether this is the best design choice when using KG-MT on other types of texts, e.g., long documents where the number of entities may be significantly higher.

\paragraph{Comparison systems.}
Our comparison systems are based on the state of the art in MT and LLMs, but they are not necessarily the best systems for the task of cross-cultural translation.
We use mBART-50, M2M-100, and NLLB-200 as the backbone for the knowledge-enhanced translator in KG-MT, as they are widely used, have shown strong performance in the literature, and are available for fine-tuning.
Another advantage is that they are also multilingual, which allows us to use the same model for all the language pairs in \benchmark{}.
Moreover, we mainly considered the GPT family of LLMs, as they are among the most popular and best performing LLMs, having also shown strong translation performance in the literature.
However, future work may consider using openly-available LLMs.
In this work, we have focused on the retrieval-augmented translation approach for MT systems, but future work may consider similar approaches for openly available LLMs, such as LLama and Mistral.

\section*{Acknowledgements}
The majority of this work has been carried out while Simone Conia and Daniel Lee were at Apple: we would like to thank all the people at Apple who provided their feedback on this work and participated in many helpful conversations. 
Simone Conia gratefully acknowledges the support of the PNRR MUR project PE0000013-FAIR, which fully funds his fellowship since October 2023.

\bibliography{main}

\appendix

\begin{figure*}[t]
	\centering
	\includegraphics[width=\textwidth]{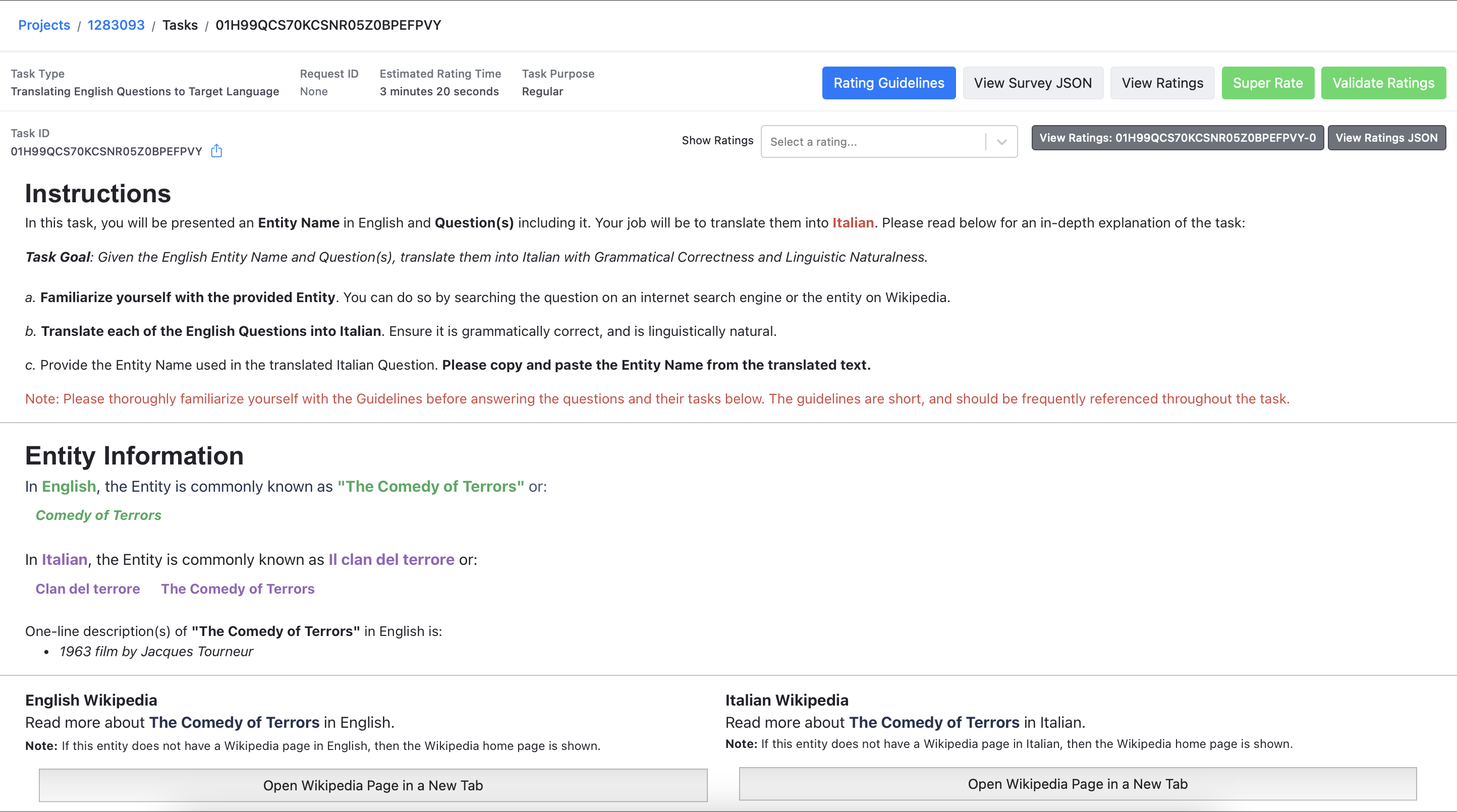}
	\caption{UI used for the annotation task: the annotators could familiarize themselves with the task with an outline of the task instructions (detailed guidelines could be read in a separate page) and the information about the entity, including its names in English and its Wikipedia pages in English and the target language (Italian in this case).}
	\label{fig:ui-first-part}
\end{figure*}

\section{Creating \benchmark{}}
In this section, we provide in-depth details on the creation process of \benchmark{}, our novel dataset for evaluating cross-cultural translation of texts containing entity names.

\subsection{Choice of languages}
As mentioned in Section 4, 10 diverse languages are selected from a set of typologically-different linguistic families:

\begin{itemize}
	\item West Germanic: German;
	\item Romance: Spanish, French, Italian;
	\item Semitic: Arabic;
	\item Sino-Tibetan: Chinese (simplified);
	\item Altaic: Turkish;
	\item Koreanic: Korean;
	\item Japonic: Japanese.
	\item Tai: Thai.
\end{itemize}

The architectural decision renders \benchmark{} a complex challenge, given the variability or consistency in the symbol sets across different languages. For instance, the spelling of a person's name might remain unchanged between English and French, yet it's highly improbable for it to be identical in English and Chinese, necessitating at least a transliteration. Furthermore, the act of transliterating between English and Korean (as well as other languages, like Japanese) is fraught with unpredictability, complicating the reliance on rule-based methods for name translation between these linguistically divergent languages. Our investigation prioritized languages classified as high/medium-resource, following the quantitative evaluation in \citet{conia-etal-2023-increasing}  indicates that the representation of textual data is substantially lacking, even for the most recognized entities (top-10\%) within those high/medium-resource languages. The extension of our benchmark to encompass lower-resource languages is earmarked for subsequent endeavors.

\subsection{Human annotation process.}
The objective of the annotation process was to (i) translate the question from English to the target language and (ii) verify the quality of the translations. This was completed through the following tasks:

\subsubsection{Translating text from English to the Target Language.} First, given the entity, the human annotators were asked to familiarize themselves with its information. Through the user interface (IU) the following details were provided: (i) entity names/aliases, (ii) short description for the given entity retrieved from Wikidata, and (iii) a built-in panel which displayed the Wikipedia article in the English and target locale, if available. By imbedding the relevant details in the UI, annotators were able to familiarize themselves with the entity without leaving the created tool. This is show in Figure 4.

After learning about the entity, the annotators were tasked with understanding the task with a set of in-depth instructions in a separate guideline.  The guidelines provided information about (i) task terminology, (2) detailed information on translating text, (3) tips and edge-cases on translating text, (4) positive and negative examples of the translation task. The guidelines will be provided in the supplementary material.

\begin{figure*}[t]
	\centering
	\includegraphics[width=\textwidth]{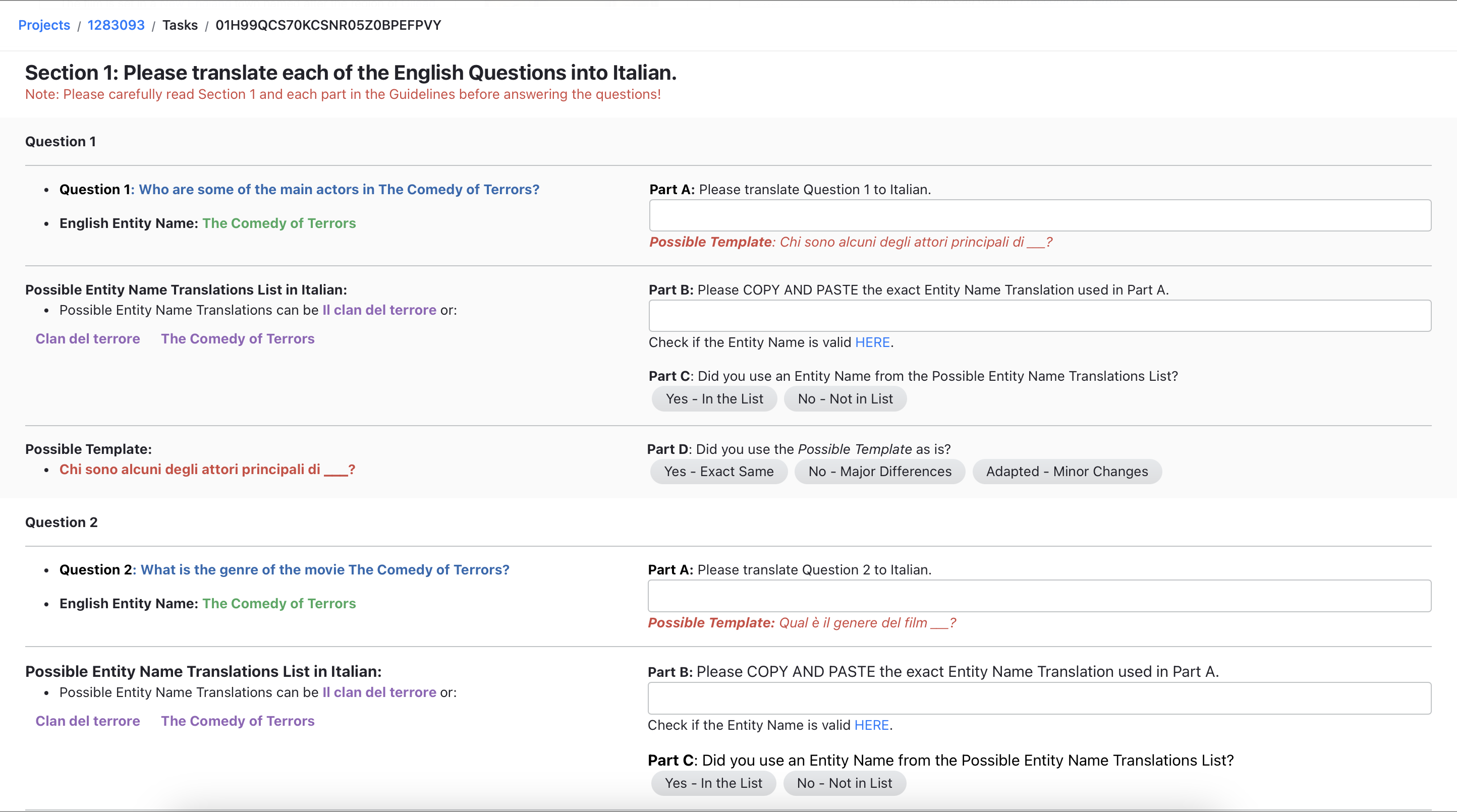}
	\caption{UI used for the annotation task: the annotator was tasked with providing the translation from the English question to the target language in a free-form text box, and was provided relevant details such as the (i) English question, (2) English entity, (3) entity names in the target language, and (4) a possible translation template in the target language.}
	\label{fig:ui-second-part}
\end{figure*}

Upon thoroughly familiarizing themselves with the task and details, the human annotators were tasked with translating upwards of 4 questions which contained the corresponding entity in a text box as shown in Figure 5. For each question, the following information was provided by the UI: (i) English question and entity name, (ii) target language, (iii) entity name in the target language, (iv) possible translation generated by a different machine translation tool.

Next,  the human annotator was requested to re-iterate the entity name in the target language in the following text box. The annotator was prompted to double check the validity of the entity name in the target language by using a Web search engine. By implementing this step, it forced the annotator to verify the validity segments of their translated text.

Upon completing the free-form componeent, the annoator was asked binary questions if the entity name in the target language they used was in the task suggested list and if the possible template for the question was used.

We note that annotators could provide feedback in case they noticed an error or had a suggested improvment. The annotation task was completed over 7 different batches, over a duration of 2 months, with iterative improvements made to the task UI and guidelines based on the feedback provided by the human annotator.

\subsubsection{Verifying human-curated translations.}
Using the data collected from the first task, a second group of human annoators were tasked with verifying the correctness of the translation.

Similar to the previous task, human annotators were provided details within the task UI and on a separate guidelines, information to familiarize themselves with the task entity, and task instructions as show in Figure 6)

After, the human annotators were tasked with verifying the translation in the target language as show in Figure 7. To do so, they were provided the (i) English entity name and question and (ii) target language entity name and question). With this information, they answered two questions, with their corresponding options:

\vspace{10pt}

\textbf{Part A} Is the Entity Name translated correctly?

\begin{itemize}
	\item \textbf{Yes} The Entity Name was translated correctly. Meaning, the translated entity name can be used to refer to the English entity. If you read the English and Translated Entity Name separately, you WOULD KNOW they refer to the same entity.
	\item \textbf{No} Meaning, the translated entity name can be used to refer to the English entity. If you read the English andTranslated Entity Name separately, you WOULD NOT KNOW they refer to the same entity.
\end{itemize}

\textbf{Part B} Does the English Question and the Translated Question have the same meaning?

\begin{itemize}
	\item \textbf{Yes} The English Question and the Translated Question DO have the same meaning. Basically, if I read the English Question and Translated Question separately, I WOULD understand the same thing.
	\item \textbf{No} The English Question and the Translated Question DO NOT have the same meaning. Basically, if I read the English Question and Translated Question separately, I WOULD NOT understand the samething.
	\item \textbf{Maybe} The English Question and the Translated Question MAYBE HAVE the same meaning. Basically, if I read the English Question and Translated Question separately, I WOULD LIKELY understand thesame thing. But, it could be interpreted differently.
\end{itemize}

The phrasing of the question in Part B was fine-tuned, to ensure annotators did not index on transliteration, and focused on the semantic meaning of the two questions.

The responses from the verification task, would be used to curate the final dataset, determining which English questions and language pair would be accepted.

\begin{figure*}[t]
	\centering
	\includegraphics[width=\textwidth]{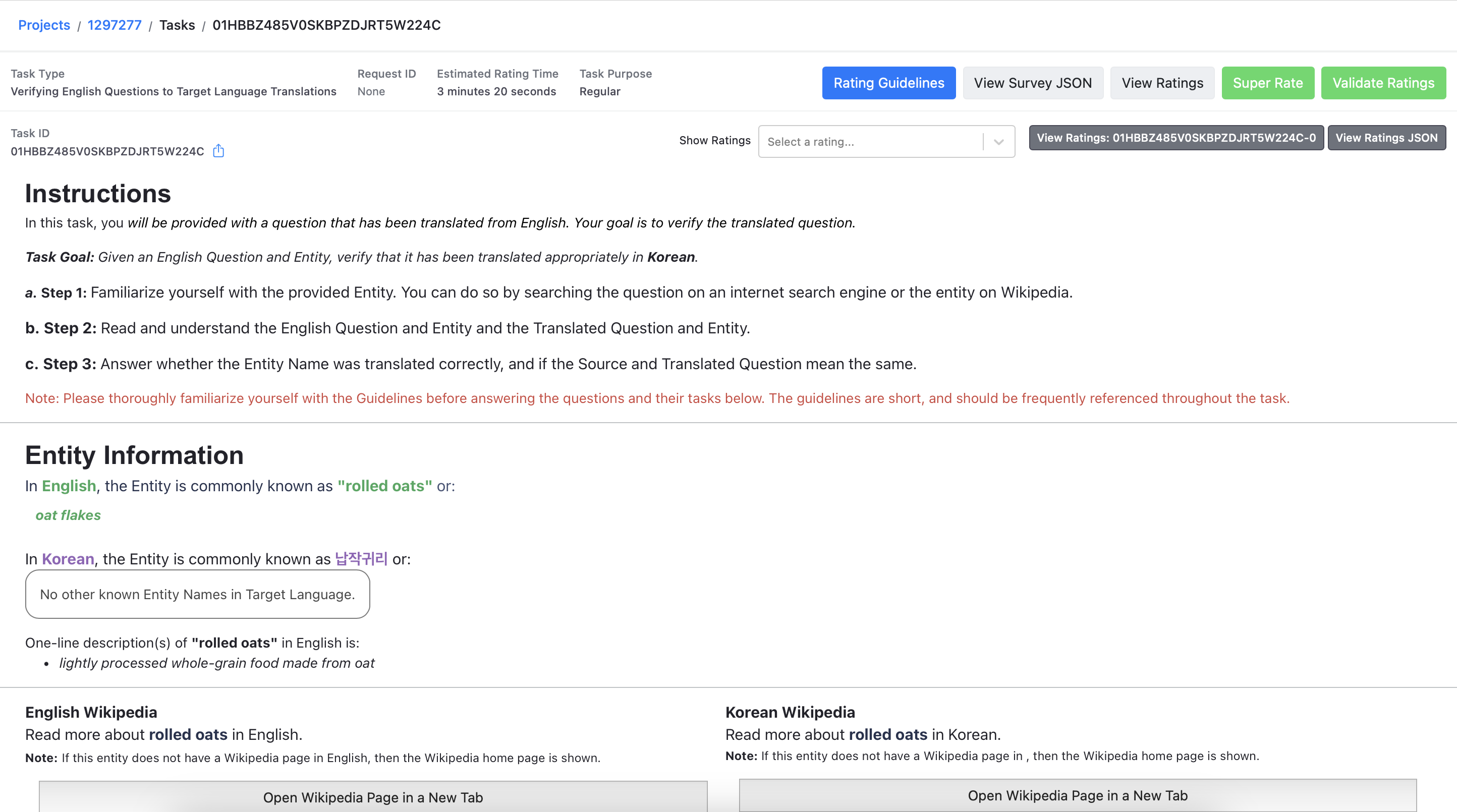}
	\caption{UI used for the annotation task: the annotators could familiarize themselves with the task with an outline of the task instructions (detailed guidelines could be read in a separate page) and the information about the entity, including its names in English and its Wikipedia pages in English and the target language (Korean in this case).}
	\label{fig:ui-third-part}
\end{figure*}

\begin{figure*}[t]
	\centering
	\includegraphics[width=\textwidth]{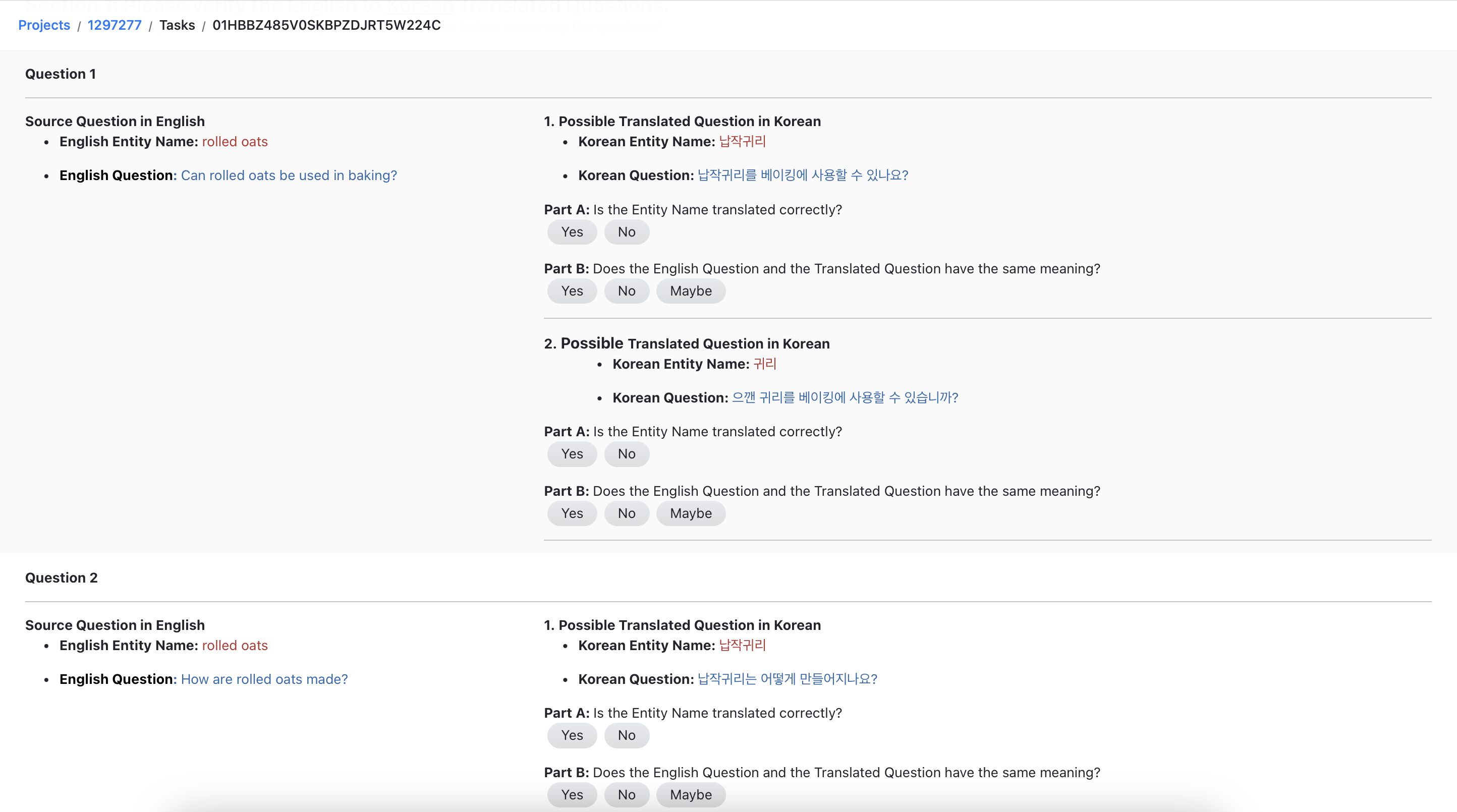}
	\caption{UI used for the annotation task: the annotator was tasked with verifying the translation from the English question to the target language in a free-form text box, and was provided relevant details such as the (i) English question, (2) English entity, (3) entity names in the target language, and (4) a possible translation template in the target language.}
	\label{fig:ui-fourth-part}
\end{figure*}

\subsection{Quality assurance and inter-annotator agreement}
To ensure the production of high-quality results, each annotator was required to clear a preliminary test before they could contribute to the annotation effort. This test involved reviewing a comprehensive guide that acquainted them with the concepts of entities and knowledge graphs, detailed the task and UI elements, and included several illustrative examples, followed by the accurate classification of 25 entity names. Those who failed the entrance exam were excluded from the actual annotation task (the 25 entities used in the test were not included in the final dataset). The pass threshold was set at 85

For each specified language, we enlisted annotators who had verified expertise in both English and the language in question, confirmed through interviews and proof of residence within the relevant country. Annotators received compensation based on the standard hourly rates for their region. On average, annotators allocated approximately one minute to acquaint themselves with the task and another minute per translation question, culminating in an average of 3.5 minutes per task. Given that each entity name was evaluated by three annotators, the cumulative human effort invested in the annotation process amounted to 3 annotators × (2800 entities × 60 seconds + 2800 entities × 3.2 questions × 50 seconds) / 60 minutes = approximately 5,133 hours.

\section{XC-Translate: Examples}
In this section, we provide examples of the instances in \benchmark{} and their
translations.

\begin{itemize}
	\item English: \textit{"Who is the author of the science fiction mystery-thriller
		      novel called The Peripheral?"}
	\item Italian: \textit{"Chi è l'autore del romanzo giallo-thriller di fantascienza
		      chiamato Inverso?"}
	\item English: \textit{"How many seasons of Sweet Magnolias are available on
		      Netflix?"}
	\item Italian: \textit{"Quante stagioni di Il colore delle magnolie sono disponibili
		      su Netflix?"}
\end{itemize}

\section{Experimental Setup}
In this section, we provide in-depth details on the experimental setup of our
experiments, including the training of the knowledge retriever, the training of
the knowledge-enhanced translator, and the evaluation metrics, as well as the
training details of the baselines.

\subsection{Hardware Infrastructure}
The experiments were conducted on a server with a single NVIDIA V100 GPU, 32GB
of RAM, and an 32-core CPU. The server runs Ubuntu 20.04 LTS and is equipped
with CUDA 12.

\paragraph{Training times.} The training of the knowledge retriever took approximately 4 hours to converge,
while the training of the knowledge-enhanced translator took approximately 6
hours to converge, depending on the underlying MT model, with M2M-100 being the
fastest and mBART-50 being the slowest. This makes our approach feasible for
training on a single GPU and for short training times contrary to synthetic
data augmentation approaches that usually require multiple GPUs and/or long
training times.

\subsection{Training of the Knowledge Retriever}
The knowledge retriever is trained using the training data from Mintaka, a
recently proposed dataset for multilingual question answering in which a subset
of the questions are tagged with the entities that appear therein. The training
data for the knowledge retriever contains instances of the form $\langle
	\mathbf{t}, e^+, e^- \rangle$, where $\mathbf{t}$ is a source text, $e^+$ is a
relevant entity for $\mathbf{t}$, and $e^-$ is an irrelevant entity for
$\mathbf{t}$, mined from Wikidata using the hard negative mining strategy
outlined in Section~\ref{subsec:retriever}.

The entire training dataset for Mintaka contains about 14,000 instances, which
makes it a relatively small dataset for training a retriever. To mitigate the
risk of overfitting, we use a pretrained retriever, mContriever, which is a
retriever trained on a large-scale multilingual dataset in a self-supervised
way.

\subsection{Training of the Knowledge-Enhanced Translator}
The knowledge-enhanced translator is trained using the training data from
Mintaka too. The training data for translation contains instances of the form
$\langle \mathbf{t}, \mathbf{t}', \hat{\mathcal{E}}_\mathbf{t} \rangle$, where
$\mathbf{t}$ is a source text, $\mathbf{t}'$ is a target text, and
$\hat{\mathcal{E}}_\mathbf{t}$ is the set of gold entities for $\mathbf{t}$.
The training data is created by uniformly sampling a mixture of the training
data from Mintaka and NLLB-200, a recent multilingual MT model that supports
translation from and to 200 different languages using a single model. While
Mintaka is a relatively small dataset, it is also convenient for our purposes,
as it can be used to train both the knowledge retriever and the
knowledge-enhanced translator. Future work may consider using larger datasets
for training the knowledge-enhanced translator as well as adopting more
sophisticated training strategies, such as curriculum learning and adversarial
training.

\subsection{Hyperparameters}
The knowledge retriever is trained using the following hyperparameters:
\begin{itemize}
	\item Learning rate: 1e-5;
	\item Batch size: 32;
	\item Number of epochs: 5;
	\item Optimizer: AdamW;
	\item Loss function: Binary Cross-Entropy;
	\item Pretrained retriever: mContriever;
	\item Hard negative mining: enabled;
	\item Number of negative samples: 8;
	\item Maximum query length: 128;
	\item Maximum context length: 128.
\end{itemize}
While the query length and the context length (i.e., the entity name and its description) are set to 128,
the textual representations of the entities usually do not exceed 100 tokens, which makes the maximum context length a reasonable choice.

The knowledge-enhanced translator is trained using the following
hyperparameters:
\begin{itemize}
	\item Learning rate: 1e-5;
	\item Batch size: 32;
	\item Number of epochs: 5;
	\item Optimizer: AdamW;
	\item Loss function: Cross-Entropy;
	\item Maximum input length: 512;
	\item Maximum output length: 512.
\end{itemize}
The maximum input length and the maximum output length are set to 512, which is the maximum length supported by the underlying MT models that we consider in our study.

\section{Related Work: Addendum}
In this section, we provide an in-depth discussion of the related work on
machine translation and entity name translation, with a particular focus on a
few more relevant works.

\paragraph{\citet{zeng-etal-2023-extract}:}
the authors recently proposed a method to extract entity names from a source text and
translate them into a target language by looking them up in a dictionary and appending their translation
to the source text, which is similar to what we named explicit knowledge integration in our work (see Section 3.2).
The authors evaluate their method on a small-scale dataset and show that it outperforms a vanilla MT system.
However, there are important differences between their work and ours:
\begin{itemize}
	\item Their focus is on transliteration rather than transcreation. This is evident in
	      their evaluation, in which they select language pairs in which transliteration
	      is necessary, but also in their method, in which they make explicit use of a
	      transliteration system. We believe that their work and ours are complementary,
	      as they focus on a different aspect of the problem, and that their method could
	      be integrated into our method to improve the performance of KG-MT.
	\item Their method is based on a dictionary lookup, which is simple and effective but
	      that ignores the problem of ambiguous entities, i.e., entities that have
	      different translations in different contexts. This is a problem that we address
	      in our work by using a knowledge retriever to retrieve relevant entities for
	      the source text.
	\item Our knowledge retriever is also capable of retrieving entities that do not have
	      an exact match with the source text, i.e., it does not rely on mention
	      detection. Moreover, using our approach allows the retriever not to retriever
	      any entities if their retrieval (or the knowledge that would be retrieved) is
	      not relevant to the translation task. In contrast, their method leaves this
	      task to the encoder-decoder architecture of the MT system.
\end{itemize}



\end{document}